\documentclass[11pt]{article}

\usepackage[utf8]{inputenc}
\usepackage[T1]{fontenc}
\usepackage{lmodern}
\usepackage[margin=1in]{geometry}
\usepackage{amsmath}
\usepackage{amssymb}
\usepackage{wasysym}      
\usepackage{graphicx}
\usepackage{booktabs}
\usepackage{tabularx}
\usepackage{array}
\usepackage{caption}
\usepackage{microtype}
\usepackage{enumitem}
\usepackage[hidelinks]{hyperref}
\usepackage{xurl}

\newcolumntype{L}[1]{>{\raggedright\arraybackslash}p{#1}}

\newcommand{\yes}{\CIRCLE}
\newcommand{\phalf}{\LEFTcircle}

\newcommand{\refitem}[1]{\par\noindent\hangindent=1.6em #1\par\vspace{2pt}}

\title{\vspace{-2em}%
\rule{\linewidth}{1pt}\\[0.55em]
{\Large\bfseries Agentic Context Management: Solving Agent Memory and Cost\\[0.15em] by Treating Them as Lifecycle and Architecture Problems}\\[0.55em]
\rule{\linewidth}{1pt}}
\author{\large Gaurav Dadhich\thanks{Maximem. Correspondence to: \texttt{gaurav@maximem.ai}.}}
\date{}

\begin{document}
\maketitle

\begin{abstract}
\noindent
Production AI agents' failures are less often because of an inability to reason well and more often because they cannot manage what is in their reasoning context; across a large number of things they must hold in context: conversation histories, large prompts, large tool definitions and ballooning tool outputs. They drown in their own accumulating history while paying for a token cost that grows with every turn, resulting in missing recalls both within the same conversation as well as across conversations. The incumbent response treats this as a storage and retrieval problem. We argue that this framing is too narrow. We propose that actively managing what an agent holds in mind is a lifecycle, not merely a store: it spans deciding what to remember, extracting and structuring it, choosing the right store for the right data-type, carving optimal redundancies, conscious consolidation, forgetting stale information while maintaining provenance, deciding what is relevant for the current turn, anticipating what will be needed next and compacting context to fit a budget without losing what matters, without compromising on recall and so on. In serious production agents, this operates not only over a single user but across an organizational scope hierarchy. We prefer naming this discipline Agentic Context Management (ACM) (as have some others in the past) and decompose it into five primitives: architecting, ingesting, scoping, anticipating, and compacting \& consolidation. We then make the economic case for why a managed lifecycle is not a luxury: na\"ive context accumulation grows token cost quadratically in conversation length, crude summarization buys linear cost at the price of an accuracy cliff and only validated compaction achieves linear cost with preserved fidelity. We describe a reference implementation, Maximem Synap, that realizes these five primitives as a multi-tenant service and reports 92\% on LongMemEval and 93.2\% on LoCoMo under the configuration detailed in Section 6. Lastly, we close with the dimensions existing benchmarks do not yet capture viz.\ latency, token efficiency and context-rot resistance and the open frontier of decision-level and organization-level context that the category points toward; but continue to determine stability and usability of an agent's utility in serious setups.
\end{abstract}

\section{Introduction}
The last two years turned LLMs from chat interfaces into agents i.e.\ systems that take actions, call tools, and carry tasks across many turns and many sessions. Multiple industry surveys underline how hard the production step is. As of 2025, a majority of enterprises are experimenting with AI agents, yet only about a quarter of them report scaling and even fewer than 10\% have scaled agents within any single business function. In fact, the large majority of agent pilots never graduate to production (McKinsey, 2025). The capability that most visibly limits this scaleup is not reasoning because frontier models reason quite well. What they rather lack is a disciplined account of what should be in the context window at each step. In the lack of such discipline, deployed agents partially or wholly forget what a user told them earlier in a long conversation, or what they fetched from tools in the same conversation, or what was mentioned to them last week; contradict themselves across a multi-agent handoff; hallucinate from a context that has been stuffed past the point of usefulness; and become more expensive with every additional turn.

The current framing for this problem is ``memory''. A healthy ecosystem of memory tools has grown up around it. We argue that the framing itself is what limits these systems. ``Memory'' names a store i.e.\ a place to put facts and get them back. A system built around a store optimizes exactly two moments viz.\ the write and the read. But let's consider the things a production agent platform must decide turn by turn: (a) which of the things just said are worth retaining at all (b) in what structure should these be retained (c) which small fraction of everything retained belongs in this turn's context (d) what is the next turn likely to need (e) what should happen when the relevant context exceeds the budget the model can use meaningfully. These are five different decisions made at five different moments, and a store makes none of them. Storage is one of the moments in a lifecycle and not the whole. A usable platform must make all these five decisions while keeping one user's context from leaking into another's and while also respecting organizational boundaries.

This paper makes four contributions that aim to address these:

\textbf{Reframing and Taxonomy:} We define Agentic Context Management and factor it into five primitives or concepts: architecting, ingesting, scoping, anticipating, compacting \& consolidation; that operate across a scope hierarchy from the individual user to the organization (Section 2).

\textbf{An economic argument:} We point out and prove that a managed lifecycle is necessary and not merely nice. A full-append context costs $O(n^2)$ in tokens over a conversation. Crude summarization trades that off for an accuracy drop. Contrarily, a validated and intelligent compaction reaches the efficient frontier of linear cost with preserved fidelity (Section 3). We support the retrieval side of the argument with an original study of retrieval across five domains of data.

\textbf{Reference implementation:} We describe Maximem Synap, a multi-tenant service that incorporates these five primitives, at the level of architecture as well as at the level of observable behavior (Sections 4--5).

\textbf{Evidence and an agenda:} We report results on two public memory benchmarks together with their limitations (Section 6), and lay out the frontier of decision-level and organization-level context management (Section 8).

Maximem's Synap product appears in this paper as the reference implementation of the category. However, the argument is about the category. Readers who finish reading this introduction section thinking in terms of context management as a lifecycle, have understood the point. The reference implementation is one way to build such a lifecycle system.

\section{From Memory to Context Management}
We define Agentic Context Management as the discipline of deciding what an agent should hold in context, when, for how long, and at what cost, across the full lifecycle from context-acquisition to context-retirement. It comprises five primitives.

\textbf{Architecting:} Before a single memory is stored, something must decide the shape of memory for a given agent: which categories of information matter, how they should be extracted, where they should live, how long they should persist, how they should be retrieved and compacted. Most systems answer this with a fixed, universal schema. We argue architecture itself is a first-class primitive to build a context-management system.

\textbf{Ingesting:} Raw signals such as conversation turns, multi-modal document uploads, tool calls and their responses; must be turned into structured, retrievable memory. The key observation that has been well established in prior work and has been confirmed in practice is that retrieval quality is bounded by ingestion quality. For example, a system that stores ``the user mentioned a pricing plan'' can never later retrieve ``the user upgraded from Starter to Pro on April 3.''

\textbf{Scoping:} At both ingestion and retrieval time, the system must decide which fraction of everything it knows is relevant for the future and at what scope. This is a layered decision across an organizational hierarchy (defined below) with strict isolation, so that one user's context never appears in another's session while one org's information grows richer to create a network-effect driven positive and productive experience for all users in a B2B agent environment; if and only if the nature of data exposed to the agent permits this behavior.

\textbf{Anticipating:} Agents don't know what they don't know. If an agent never asks for it, they are unlikely to retrieve it. Speculative prefetching is a long-standing idea in computer systems. We have applied it to the agent context. It enables a context-management system to in-principle observe an agent's behavior and prepare the context it will likely need before the explicit request arrives, moving retrieval off the critical path. We treat anticipation as a primitive distinct from retrieval: retrieval answers ``what is relevant to a search query or the moment,'' anticipatory retrieval answers ``what will be relevant next''. It also helps an agent get answers to questions that it doesn't know it needs to know, to be able to serve a user requirement accurately.

\textbf{Compacting \& Consolidation:} When the relevant context exceeds the budget that the downstream model can use well, the system must reduce it; without discarding what will be needed. Crucially, compaction should be verifiable: a compaction that silently drops a critical fact is worse than no compaction, because it produces confident wrong answers. We propose that verifiable, lossless and opinionated compaction is possible and valuable at production-scale.

These five primitives are coupled: the architecture chosen by the first changes what the others should do (a support agent and a coding agent need different categories, retention, and compaction). That coupling is the core argument for managing context as a system rather than assembling five independent tools.

\textbf{The scope dimension:} Each primitive operates across a scope hierarchy, not just per-user. Throughout this paper we use three terms for the hierarchy, from narrowest to broadest: the user (an individual person or agent / sub-agent interacting with an agent), the customer (the organization that user belongs to) and the client (the operator of the agent platform, whose deployment spans many customer organizations). Both Ingestion \& Retrieval should resolve this hierarchy narrowest-first i.e.\ user, then customer, then client; under strict isolation. A separate, global knowledge layer should exist for shared general knowledge (for example, canonical identities of public entities). Most memory tools scope to the user; flattening organizational structure into a single bucket either loses organizational context or leaks one user's data into another's session. Treating scope as a first-class dimension: primitives $\times$ scopes: is what lets context management serve organizations rather than only individuals.

\begin{figure}[htbp]
\centering
\includegraphics[width=\textwidth]{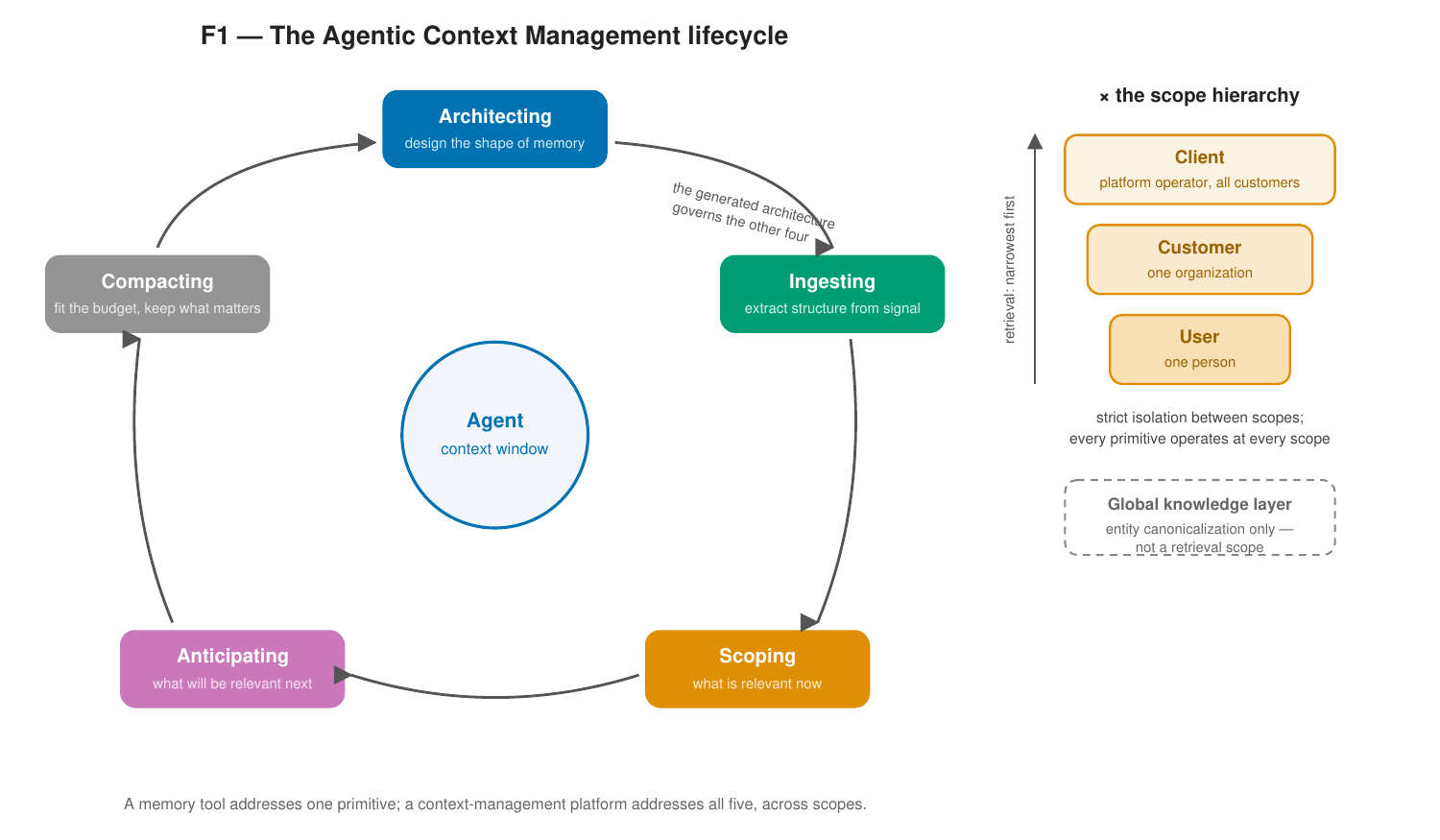}
\caption{The five-primitive context lifecycle (architecting $\rightarrow$ ingesting $\rightarrow$ scoping $\rightarrow$ anticipating $\rightarrow$ compacting \& consolidation), drawn as a cycle around a central agent, with the retrieval scope hierarchy (user $\rightarrow$ customer $\rightarrow$ client) as a vertical axis and the global knowledge layer drawn separately, feeding entity canonicalization.}
\label{fig:lifecycle}
\end{figure}

We propose that a system that addresses one primitive well is a memory tool. A system that addresses all five coherently, across scopes, is a context-management platform.

The taxonomy is grounded in observed failures: Each primitive earns its place because its absence has a named, observed production failure mode; several documented in our own deployment notes (Maximem, 2026a), others in the published literature. Table~\ref{tab:failures} makes the mapping explicit; Section 3 quantifies the two failure modes that dominate cost and accuracy.

\begin{table}[htbp]
\footnotesize
\centering
\caption{Production failure modes and the primitive whose absence produces them}
\label{tab:failures}
\begin{tabularx}{\textwidth}{L{2.4cm} X L{2.8cm} L{2.3cm}}
\toprule
\textbf{Observed failure} & \textbf{What it looks like in production} & \textbf{Absent primitive} & \textbf{Evidence} \\
\midrule
Junk accumulation & Low-value or duplicate entries accumulate and crowd out useful memory. An audit of one popular memory library found 10,134 entries stored over 32 days, of which 38 were usable, a 99.6\% junk rate (boot-file restatements, cron noise, config dumps) & Architecting + Ingesting (store-first, extract-later) & Maximem (2026a)\textsuperscript{g} \\
Lost detail & Extraction keeps a vague paraphrase, so the specific fact can never be retrieved. ``User mentioned a plan'' stored where ``user upgraded from Starter to Pro on April 3'' was said; no retrieval method can recover the detail that extraction discarded & Ingesting & Maximem (2026a); Section 3.3 \\
Identity fragmentation & One real entity is stored under several unlinked names, splitting its history. ``Sarah,'' ``Sarah Chen,'' and ``SC'' stored as three unrelated strings; retrieval returns whichever matches the query embedding, yielding incomplete or contradictory answers & Ingesting (entity resolution) & Maximem (2026a) \\
Scope bleeding & Memory from one scope appears in another that should be isolated from it. One user's preferences surface in another user's session; or the agent misses organizational context entirely & Scoping & Maximem (2026a) \\
Cross-session amnesia (surfacing as handoff repetition) & Knowledge from earlier sessions is not carried forward, so it must be re-established. The user repeats themselves to every new session or every sub-agent & Scoping (lifecycle absent) & Maximem (2026a) \\
Retrieval on the critical path & Every turn blocks on a synchronous retrieval round-trip before the model can respond. Every turn stalls on a retrieval round-trip & Anticipating & Section 5 \\
The accuracy cliff & Unvalidated compression removes information needed later, collapsing accuracy. 18,282 tokens compressed to 122 in one unvalidated step; accuracy fell from 66.7\% to 57.1\%, below the no-context baseline & Compacting \& Consolidation (validation absent) & Zhang et al.\ (2025) \\
Quadratic cost growth & Re-sending an unmanaged, growing context makes token cost rise quadratically. Token cost per conversation grows with the square of its length & Compacting \& Consolidation & Section 3.1 \\
\bottomrule
\end{tabularx}
\end{table}

\section{Why Store-and-Retrieve Is Not Enough}
The reframing in Section 2 is conceptual. This section makes it quantitative, because the case for a managed lifecycle is ultimately an economic and an information one, and both are measurable.

\subsection{The cost of doing nothing grows quadratically}
Consider a multi-turn agent conversation. Let each turn add about $t$ tokens (user message plus assistant reply), and let the conversation run $n$ turns. In the na\"ive ``full-append'' pattern (re-send the entire history every turn, which is what most hand-rolled agents do), the input context at turn $k$ is the whole history so far, roughly $k\!\cdot\!t$ tokens. Cumulative input tokens across the conversation are therefore
\[
C_{\text{append}} \;=\; \sum_{k=1}^{n} k\,t \;=\; t\,\frac{n(n+1)}{2} \;\approx\; \frac{t}{2}\,n^2 \;=\; O(n^2).
\]
Because providers bill per input token, cost grows quadratically with conversation length. (Per-call attention compute is worse, quadratic in sequence length per turn, but the billing result is the cleaner one to reason about.) If instead the system holds context to a fixed budget $W$ each turn, cumulative tokens are $n\!\cdot\!W = O(n)$. The ratio,
\[
\frac{C_{\text{append}}}{C_{\text{bounded}}} \;=\; \frac{t\,(n+1)}{2W},
\]
grows linearly with $n$: the longer the conversation, the more punishing full-append gets. For illustrative values ($t = 500$, $W = 4{,}000$), the multiple is roughly $6\times$ at 100 turns and $13\times$ at 200 turns; the full derivation and a sensitivity table are in Appendix A.

\begin{figure}[htbp]
\centering
\includegraphics[width=0.86\textwidth]{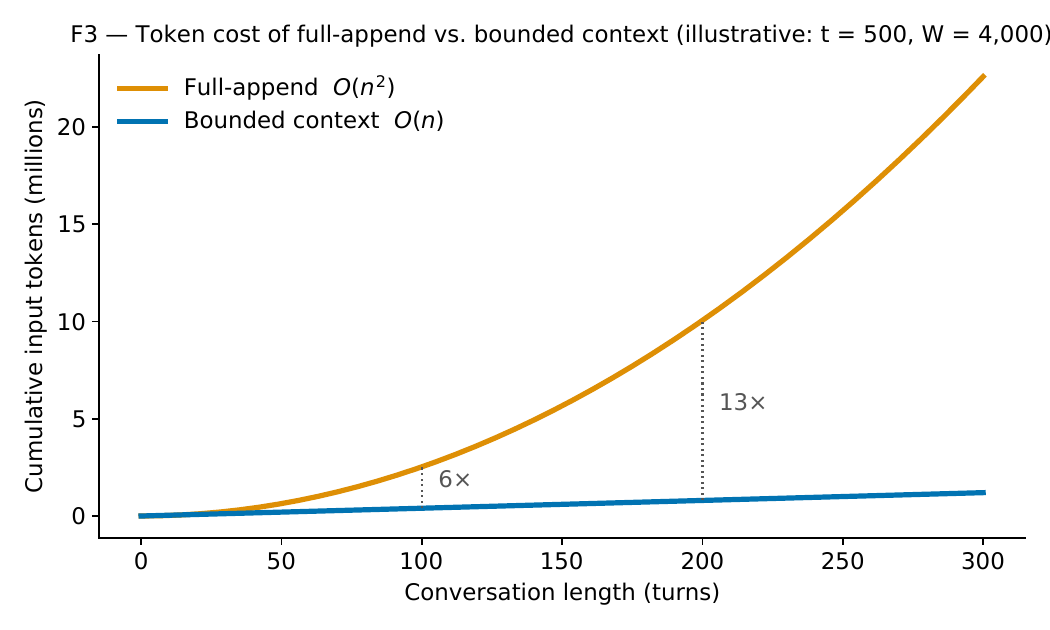}
\caption{Cumulative input tokens vs.\ turns: full-append (quadratic) against bounded context (linear).}
\label{fig:tokencost}
\end{figure}

\subsection{But bounding context na\"ively destroys accuracy}
Bounding the budget is necessary; how you bound it decides whether you keep accuracy. Crude summarization bounds tokens but is lossy and unvalidated, and the loss can be catastrophic: prior work documents a case in which compressing an 18,282-token context to 122 tokens in a single step dropped task accuracy from 66.7\% to 57.1\%, worse than providing no context at all (Zhang et al., 2025). The summarizer threw away what mattered because it had no way to know what would be needed downstream.

This yields a three-way comparison:

\begin{center}
\small
\begin{tabularx}{\textwidth}{L{3.2cm} L{2.2cm} L{3.0cm} X}
\toprule
\textbf{Approach} & \textbf{Token cost} & \textbf{Fidelity} & \textbf{Failure mode} \\
\midrule
Full-append & $O(n^2)$ & Full, until context rot & Cost explodes; long contexts degrade (``lost in the middle,'' Liu et al., 2024) \\
Crude summarization & $O(n)$ & Lossy, unvalidated & The accuracy cliff \\
Validated compaction & $O(n)$ & Preserved + checked & None (the target) \\
\bottomrule
\end{tabularx}
\end{center}

\begin{figure}[htbp]
\centering
\includegraphics[width=0.78\textwidth]{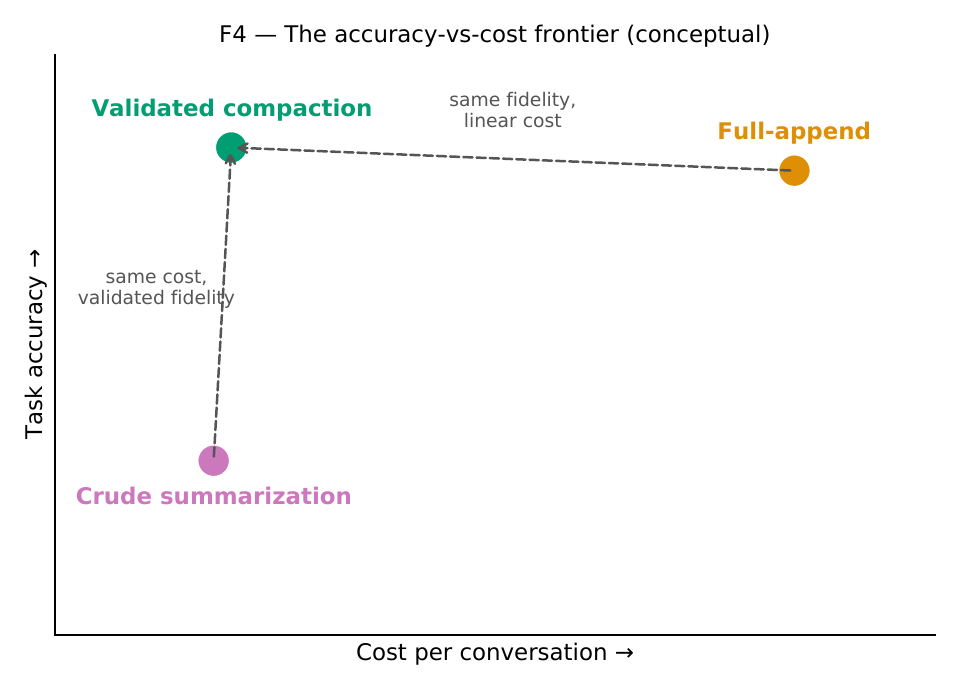}
\caption{The accuracy-vs-cost frontier: full-append (top-right), crude summarization (bottom-left), validated compaction (top-left). One figure that carries the section.}
\label{fig:frontier}
\end{figure}

The argument is now sharp. A context-management system should sit in the top-left corner, linear cost with validated fidelity, and getting there requires treating compaction as a verified operation rather than a hopeful one.

\subsection{Retrieval is not the same as sufficiency}
Cost is one half the story. The other half is whether retrieved context is enough to reason with. Retrieval quality is itself a chain of bottlenecks:
\[
\text{answer quality} \;\le\; \min(\text{extraction quality},\ \text{retrieval quality},\ \text{reasoning sufficiency}).
\]
Extraction here are the ingestion-time steps that turn raw signal into stored, structured memory. Retrieval are the query-time steps that fetch these back. Each of the steps cap the output-answer quality. If you store junk, the extraction caps you. The junk-rate audit of Table~\ref{tab:failures} is the extreme case, where faithful storage of unextracted signals made the store effectively unusable. Retrieve the wrong material and retrieval caps you; retrieve some of the relevant material but not all the context needed to reason to a provable answer, and reasoning sufficiency caps you (Dadhich, 2026a). The last is the most overlooked, most benchmarks measure a retrieval hit (did a relevant document appear?), almost none measure whether everything required to reason appeared, let alone how many irrelevant items were retrieved.

We ran a motivating study (purely out of curiosity) of retrieval across five domains: code (CodeXGLUE), web (MS MARCO), facts (SQuAD), multi-hop reasoning (HotpotQA), and science (SciQ): five corpora of 10,000 documents each, 1,000 queries per corpus, scored by MRR@10 (Dadhich, 2026a; data and per-dataset results at maximem-ai/file-vs-vector-study-results). Let us begin by stating its limitations: (a) single operator, one keyword engine against one vector store (b) no chunking (c) small per-corpus scale. We present it as motivation, not as a controlled benchmark (full methodology and caveats in Appendix B). But the two findings that are relevant here. (a) vector and keyword retrieval win in different regimes, and not with a narrow margin. Vector search dominates where the semantic gap is widest (natural-language-to-code: 0.91 vs.\ 0.29 MRR, a query for ``sort a list'' finding bubble\_sort), while keyword search wins decisively wherever terms are specific entities rather than vibes or semantics (science QA: 0.81 vs.\ 0.61, ``mitochondria'' is a key, not a similar concept), with factoid QA at parity and multi-hop slightly keyword-leaning (Table B1). The study also quantified the vector tax: indexing the same 10,000-document corpus took 60--100$\times$ longer with embedding generation than with keyword indexing (Table B2), a real constraint when an agent must read new material now and act on it now.

However, the deeper finding is what the study was not able to measure, by design. Like most retrieval evaluations, it scores a hit when a single gold document appears in the top results such as on HotpotQA it scores against one supporting document per question, even though multi-hop questions require two or more. An evaluation built this way cannot detect the failure that matters most for agents i.e.\ retrieving a relevant document while missing the bridge document needed to complete the reasoning chain. It is also an effect we observed qualitatively while inspecting outputs but given the single-target scoring could not quantify, and which we therefore report as an observation rather than a result. This is the reasoning-sufficiency gap in its purest form wherein retrieval hits are measurable everywhere; reasoning sufficiency, almost nowhere. The conclusion we draw is that retrieval must combine lexical and semantic signals: a hybrid of keyword and vector search, since each covers the other's blind spot. With our experiments in building Maximem's Vity (a personal AI memory vault across apps), we learnt that vector-only implementations have their own limitations and need to be used in conjunction with graphs so as to retain relational information in querying as well as to manage provenance. So the graph gets added to the mix. But this hybrid approach alone is necessary, not sufficient. Closing the sufficiency gap also needs structured ingestion and scope-aware assembly, which is why the rest of the paper treats retrieval as one primitive in a managed lifecycle rather than a standalone component, and why Section 6.3 argues the field needs evaluations that score sufficiency directly.

\begin{figure}[htbp]
\centering
\includegraphics[width=\textwidth]{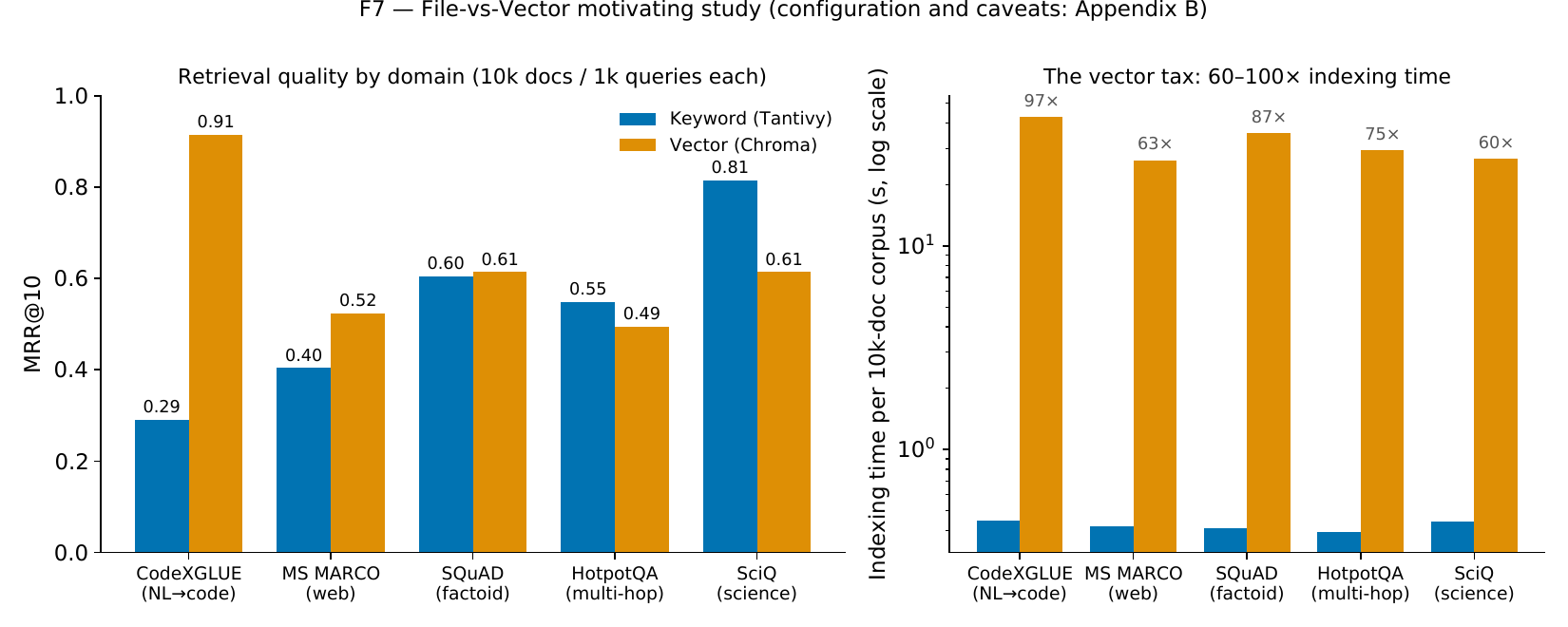}
\caption{MRR by dataset for keyword vs.\ vector retrieval, with the embedding-generation (``vector tax'') latency overlaid.}
\label{fig:mrr}
\end{figure}

\subsection{What the lifecycle has to provide}
The two arguments land to the same conclusion. Closing the cost gap requires validated compaction; closing the sufficiency gap requires ingestion that preserves structure, scoping that assembles sufficient context, and retrieval that combines semantic and relational signals. No single retrieval method delivers this but a managed, purpose-designed lifecycle does. The next section describes one such system.

\section{The Maximem Synap System}
Maximem Synap is a multi-tenant, hosted context-management service. Throughout this section we describe components by what they do and why: their interfaces, observable behavior, and guarantees, and deliberately not by how they work internally; the mechanism internals are proprietary. This single statement covers the whole section; we do not annotate individual components.

\begin{figure}[htbp]
\centering
\includegraphics[width=\textwidth]{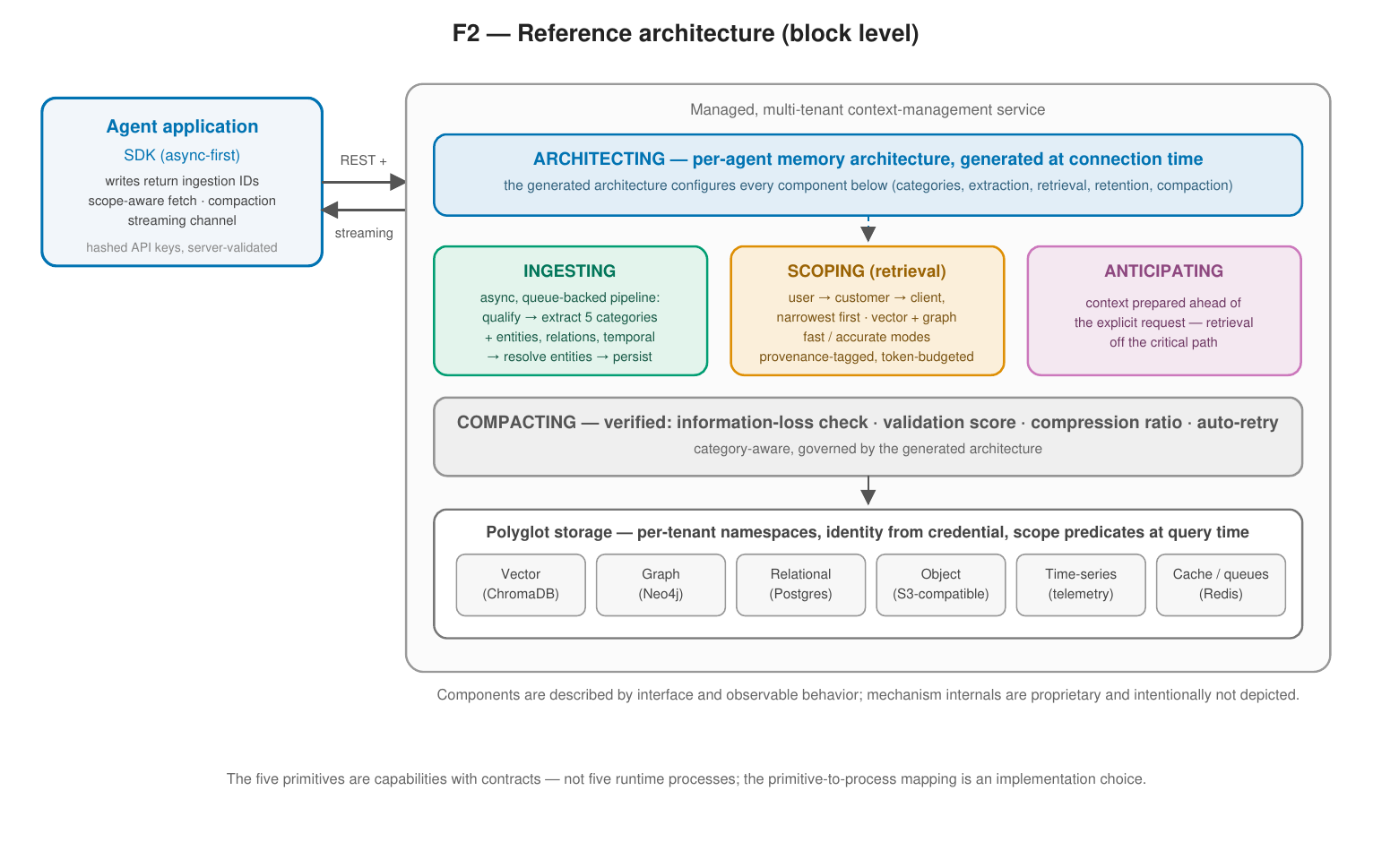}
\caption{Reference architecture: SDK $\rightarrow$ API (REST + streaming) $\rightarrow$ managers/pipelines $\rightarrow$ polyglot storage, with the five primitives annotated onto the components that realize them.}
\label{fig:arch}
\end{figure}

\textbf{Client surface:} Maximem Synap exposes an async-first SDK (Python canonical; a JavaScript bridge) whose memory-write calls return an ingestion identifier immediately and never block the calling application; processing happens asynchronously. The SDK offers a unified, scope-aware retrieval call, a conversation-compaction call, and a streaming channel. Tenant isolation is enforced at both the storage layer (per-tenant namespaces) and the query layer (scope predicates), with identity derived from the stored credential rather than asserted by the client.

\textbf{Architecting (per-agent memory design):} When an agent is connected, Maximem Synap generates a bespoke memory architecture for that agent from the customer's description of its purpose and any reference material it is given, choosing which memory categories to capture, and how to extract, store, retrieve, and compact them, and then activates that architecture. This is an autonomous, LLM-reasoned design step with multi-agent checks and validations, not selection from a fixed menu; the resulting architecture governs the behavior of every downstream primitive.

\textbf{Ingesting:} Ingestion is an asynchronous, queue-backed multi-stage pipeline. It accepts a document, returns an ingestion ID immediately, and in the background qualifies the content, extracts memory categories: facts, preferences, episodes, emotions, temporal events, and more depending upon the nature of agents the system is supporting in that instance; together with entities, relationships, and temporal validity, then resolves entities and persists the results across relational, graph, and vector stores. Extraction decisions are governed by the per-agent architecture from the setup step and not based on a fixed universal schema.

\textbf{Entity resolution:} Because the same person or thing appears under many surface forms (``Sarah,'' ``Sarah Chen,'' ``SC''; or, at the organizational level, ``PR FAQ'' and ``6-pager'' for one document), Maximem Synap resolves extracted entities to canonical identities during ingestion. It runs a confidence-ordered cascade of matching strategies, from exact identifiers through progressively softer lexical, semantic, and contextual signals. Public entities are checked against the global knowledge layer. Anything it has not seen before is registered at the customer scope. Resolution is best-effort and never blocks ingestion; ambiguous matches can be queued for review.

\textbf{Scoping (retrieval):} Retrieval is a scope-aware pipeline that resolves the hierarchy narrowest-first: user, then customer, then client, and returns provenance-tagged, token-budgeted, ranked memory items. It offers a low-latency mode and a higher-accuracy mode that adds query decomposition and LLM reranking.

\textbf{Graph-aware retrieval:} Beyond vector similarity, Maximem Synap performs vector-guided multi-hop traversal of a knowledge graph: semantic similarity informs where to enter the graph and which relationships are worth following, surfacing related memories that pure vector search would miss (the bridge-document problem of Section 3.3). Graph-augmented retrieval has published precedents (Edge et al., 2024; Guti\'errez et al., 2024; Hu et al., 2025), which we discuss in Section 7; Maximem Synap's specific traversal and scoring approach is not described here.

\textbf{Anticipating:} Maximem Synap additionally implements an anticipatory retrieval path: context an agent is likely to need can be prepared ahead of the explicit request, with the design intent of both telling the agent what it needs but doesn't know exists in memory and also of moving retrieval latency off the agent's critical path, and with anticipatory and explicit retrieval governed by the same decision logic. This is not a query-result cache because a cache replays answers to repeated queries. Anticipation on the other hand predicts from the agent's evolving behavior; it anticipates and pre-fetches the context that has not yet been requested and prepares it before the request arrives. Its value is latency rather than deduplication since retrieval sits on the agent's critical path (Section 5). Moving the expected fetch off that path turns a blocking round-trip into a cache read on the turn that needs it and at the cost of speculative work that is discarded on a miss. We have described the concept but not the prediction mechanism which in itself is proprietary and took us a while to achieve a high-hit rate while minimizing wasteful work that the system needs to perform to reduce compute and intelligence footprint and is also an active work in progress (currently we achieve a 60\%+ hit-rate consistently across clients).

\textbf{Compacting \& Consolidation:} For long conversations, Maximem Synap treats compaction as a verified operation: each compaction is checked for information loss. The system tests whether key information from the original conversation remains recoverable from the compacted result, and emits an explicit validation score and compression ratio, automatically retrying with less aggressive compression when validation falls below threshold. Compaction is category-aware: what must be preserved verbatim and what may be abstracted is governed by the agent's generated architecture. This is the validated compaction of Section 3.2; the validation mechanism itself is not described here.

\textbf{Storage:} Maximem Synap uses a polyglot, per-tenant-namespaced stack: a vector store for embeddings, a graph store for the entity--relationship graph, a relational store as source of truth and for short-term context, an object store for original content and per-agent configurations, a time-series store for usage telemetry, and an in-memory store for queues, cache, and coordination. The specific engines behind each store are implementation choices and may change or be combined; what matters at this level is the role each store plays. The vector and graph stores together are what let retrieval combine semantic and relational signals, the hybrid that Section 3.3 argued is necessary: the vector store bridges the semantic gap, while the graph store supplies the relational links a pure similarity search misses. Embeddings are computed with locally hosted sentence-transformer models by default.

\textbf{The integration pattern:} From the application's side, the lifecycle reduces to a three-call pattern around each model invocation: scoped retrieval of past context, validated compaction of the current conversation, and non-blocking ingestion of the new turn (Listing 1, in pseudocode; concrete signatures are in the public SDK documentation). The point of the listing is what is absent from it: no schema design, no embedding-model choice, no index management, no isolation logic. Those are the lifecycle's job.

\begin{quote}
\ttfamily\footnotesize\raggedright
Listing 1 --- production integration pattern (pseudocode), on each turn:\\[3pt]
retrieved $\leftarrow$ FETCH(query=user\_message, scope=\{user, customer\})\ \ $\triangleright$ scoped retrieval\\
compacted $\leftarrow$ COMPACT(current\_conversation)\ \ $\triangleright$ validated compaction\\
reply\ \ \ \ \ $\leftarrow$ MODEL(assemble(retrieved, compacted, recent\_turns))\\
INGEST(turn, scope)\ \ $\triangleright$ asynchronous; returns an ingestion id immediately
\end{quote}

\textbf{An example trace:} Consider a support agent for a SaaS product receiving: ``Hi, Sarah said I should ask you. We upgraded to Pro last week but the dashboard still shows Starter.'' At connection time, architecting has already produced this agent's memory architecture (billing-relevant categories, customer-scoped organizational entities, conservative compaction for account facts). Ingestion extracts a fact with temporal validity (upgraded Starter $\rightarrow$ Pro, last week), an episode (dashboard shows stale plan), and an entity mention (``Sarah''), which entity resolution links to a canonical identity already known from earlier mentions of ``Sarah Chen'' and ``SC'' at this customer's scope. The same first name at a different customer resolves to a different person. On the next turn, scoped retrieval assembles user-level context (this person's open tickets), customer-level context (the organization's plan history and team), and client-level patterns (a known plan-propagation delay), each item provenance-tagged and fitted to the token budget. As the conversation grows long, compaction reduces it and reports a validation score before the next model call. The application code is Listing 1 throughout; every behavior in this paragraph is observable through the SDK and dashboard.

\textbf{A note on decomposition:} We present Maximem Synap as five primitives, capabilities with defined contracts, rather than five independent runtime components. The mapping from primitives to processes is an implementation choice, and it has changed over the system's life. The contracts are the stable surface.

\section{Design Choices}
We highlight the choices that distinguish a context-management system from a memory tool, each tied to a failure mode from Section 3.

\textbf{Architecture is synthesized and custom, not fixed and generic:} A universal schema cannot serve a fintech support agent and a coding assistant equally; their categories, retention, and compaction needs differ. Generating the architecture per agent is what makes the rest of the lifecycle fit the use case, and it is the choice most responsible for extraction quality (the first link in the Section 3.3 chain). It is also the choice that makes the coupling argument of Section 2 concrete: the generated architecture is the artifact through which the architecting primitive configures the other four.

\textbf{Compaction carries a quality contract:} Because an unvalidated compaction produces confident wrong answers (Section 3.2), Maximem Synap returns an explicit validation score and compression ratio with every compaction and retries automatically when validation fails. The system provides a confirmation of whether the compression worked instead of compressing and hoping that it did the job well.

\textbf{Validation is not free:} compacting a context and checking it spends tokens. But compaction is not a one-time event; it runs periodically, each pass operating on the already-compacted context plus recent turns rather than on the full transcript. Every compaction therefore compresses a bounded context, and the number of compactions grows only linearly with the conversation, so the overhead is linear rather than quadratic. If the context is held near a budget $W$ and compaction fires every $p$ turns at a cost of $c$ times the bounded context, the total token cost over $N$ turns is $N\!\cdot\!W\!\cdot\!(1 + c/p)$: the linear cost of Section 3.1 raised by a fixed factor $(1 + c/p)$. Measured against the $O(N^2)$ full-append baseline, the saving grows with conversation length rather than shrinking; at $t = 500$ tokens per turn, $W = 4{,}000$, $p = 8$, and $c = 2$ (a fixed $1.25\times$ overhead), net token savings run to roughly 80\% at 100 turns, 90\% at 200, and 96\% at 500. This repeated re-compaction is safe only because each pass is validated: without the information-loss check, iterated compression drifts into the context-collapse failure of Section 3.2.

\textbf{Ingestion is asynchronous and non-blocking:} Returning an ingestion ID immediately keeps the calling agent responsive; the heavy work (extraction, resolution, multi-store persistence) happens off the critical path. That trade of immediate consistency for latency is deliberate. It works because memory is read far more often than it is written. This does not sacrifice read-your-writes within a session: the most recent turns are carried verbatim in the working context (the recent\_turns of Listing 1), so information the agent just produced or received is available on the next turn without waiting on ingestion. Asynchrony defers only the durable, structured, long-term availability of that information, not its immediate use.

\textbf{Scope is first-class and isolation is enforced, not advised:} Multi-tenant isolation is implemented at the storage and query layers with identity derived from the credential, so organizational context is available and one user's data cannot surface in another's session, the failure that flat, user-scoped memory invites. Scope policy is Maximem-advised but client-governed: Maximem proposes the scoping an agent should use, and the client approves and gates it, so the customer retains control over what is shared across users and organizations and what stays isolated.

\textbf{Retrieval combines semantic and relational signals:} Pure vector search misses bridge context (Section 3.3). Pure graph traversal is expensive and brittle. Combining the two aims at reasoning sufficiency, not just at retrieval hits.

\textbf{Latency is designed around, not just optimized:} Retrieval sits on an agent's critical path, so the system offers an explicitly low-latency retrieval mode and, beyond that, the anticipatory path of Section 4 exists to take retrieval off the critical path altogether when context can be prepared in advance.

\section{Evaluation}
\subsection{Setup}
We evaluate Maximem Synap on two public, third-party benchmarks of conversational memory: LongMemEval (Wu et al., 2024), 500 human-curated questions over long multi-session chat histories probing six ability categories, and LoCoMo (Maharana et al., 2024), question answering over very long-form conversations ($\sim$300 turns on average). Neither benchmark was designed by us; both are used in their official public distributions, with no custom subsets and no relabeling.

Table~\ref{tab:config} specifies the exact configuration for each reported result. We consider an evaluation number interpretable only when its full configuration is stated, and we hold our own numbers to that standard. One scope decision deserves emphasis: LoCoMo's category 5 is adversarial: questions designed to be unanswerable, measuring abstention rather than memory, and its inclusion or exclusion moves the headline score by ten points or more, which is the most common source of incomparable LoCoMo numbers. We report categories 1--4 and exclude category 5, matching the convention of the original paper, Mem0, and Zep (Maximem, 2026d).

\begin{table}[htbp]
\footnotesize
\centering
\caption{Evaluation configuration (full methodology: Maximem, 2026d)}
\label{tab:config}
\begin{tabularx}{\textwidth}{L{2.6cm} X X}
\toprule
 & \textbf{LongMemEval} & \textbf{LoCoMo} \\
\midrule
Result (overall) & 92.0\% (460 / 500) & 93.2\% (categories 1--4) \\
Dataset & LongMemEval\_S, full 500-question set, all 6 categories, official distribution & locomo10, official distribution; Cat 1--4 (adversarial Cat 5 excluded per convention) \\
Answer model & gpt-5-mini & gpt-5-mini \\
Judge & gpt-5-mini, binary CORRECT/WRONG against gold & gpt-5-mini, binary CORRECT/WRONG against gold \\
Retrieval configuration & Scope-aware retrieval over the ingested memory; full per-run configuration in the released methodology & same \\
Harness & {\urlstyle{same}\nolinkurl{maximem-ai/memory_and_context_eval_harness}} (Maximem, 2026e) (open) & same \\
Artifacts & Methodology + per-category counts public (Maximem, 2026d); per-run artifacts (answers, retrieved context, judge verdicts) available on request & same \\
Run date / repo tag & Tagged in the results repository ({\urlstyle{same}\nolinkurl{maximem-ai/eval_benchmark_runs_output}}) & same \\
\bottomrule
\end{tabularx}
\end{table}

\subsection{Results}
Under the configuration in Table~\ref{tab:config}, Maximem Synap reaches 92.0\% overall on LongMemEval (460/500) and 93.2\% on LoCoMo categories 1--4, consistent with our public record (Maximem, 2026c; 2026d). Per-category results follow; we report the weak categories as plainly as the strong ones.

\begin{center}
\small
\textbf{Per-category results} (official category distributions; counts in Maximem, 2026d)\\[4pt]
\begin{tabularx}{\textwidth}{L{4.2cm} L{3.0cm} L{3.6cm} L{2.2cm}}
\toprule
\textbf{LongMemEval category} & \textbf{Score (correct/n)} & \textbf{LoCoMo category} & \textbf{Score} \\
\midrule
Single-session-user & 100.0\% (70/70) & Multi-hop & 97.3\% \\
Single-session-preference & 100.0\% (30/30) & Open-domain & 93.4\% \\
Knowledge-update & 100.0\% (78/78) & Temporal & 90.8\% \\
Temporal-reasoning & 100.0\% (133/133) & Single-hop & 88.8\% \\
Single-session-assistant & 87.5\% (49/56) & Overall (Cat 1--4) & 93.2\% \\
Multi-session & 75.2\% (100/133) & & \\
Overall & 92.0\% (460/500) & & \\
\bottomrule
\end{tabularx}
\end{center}

Residual errors concentrate in LongMemEval's multi-session category (75.2\%): reasoning that must join information across separate sessions, and, to a lesser degree, single-session-assistant (87.5\%). We report this plainly: multi-session reasoning is the hardest published category for every system we are aware of, and it is precisely the reasoning-sufficiency regime of Section 3.3.

\begin{figure}[htbp]
\centering
\includegraphics[width=\textwidth]{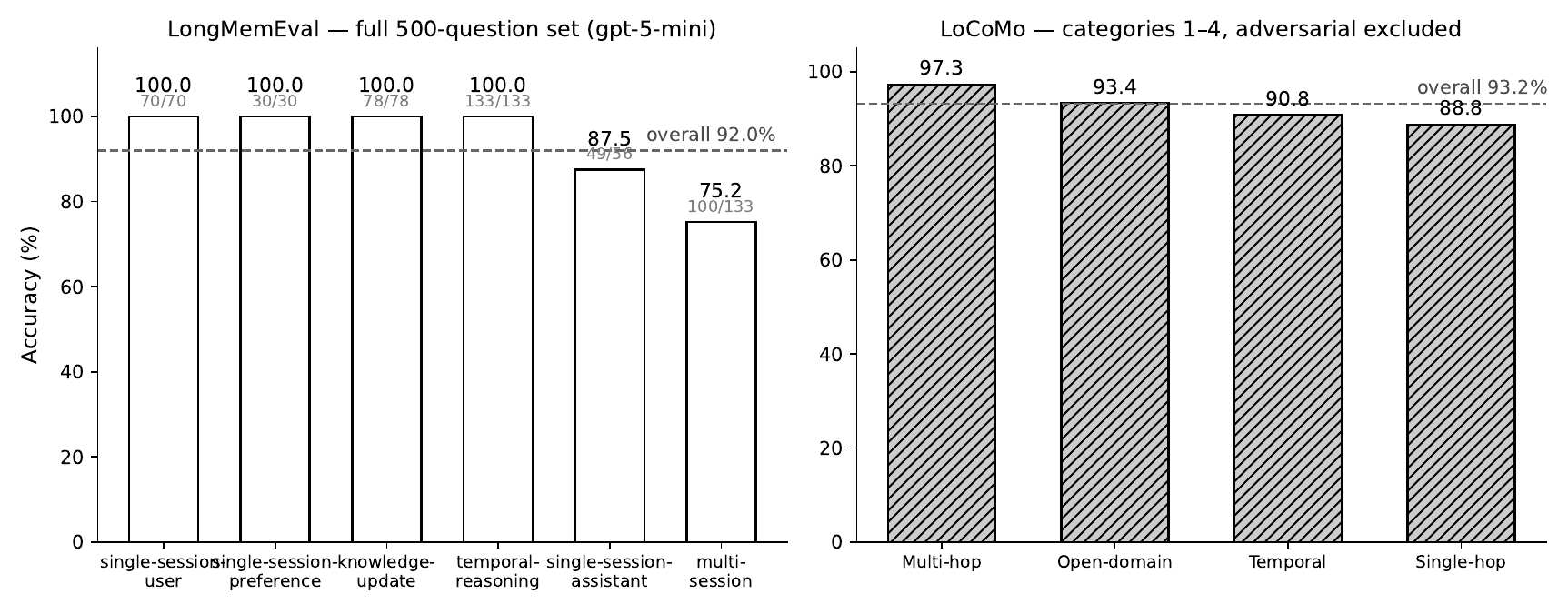}
\caption{Per-category results: LongMemEval six categories (92.0\% overall) and LoCoMo categories 1--4 (93.2\% overall).}
\label{fig:percat}
\end{figure}

\textbf{In comparison.} LoCoMo in particular has been the subject of public methodology disputes between vendors, and memory-benchmark scores are highly sensitive to answer model, judge, ingestion granularity, and adversarial-category handling; the same system can move several points on the answer model alone (SuperMemory's own published sweep spans 81.6\%--85.2\% across answer models). We therefore keep two rules. First, our own configuration is stated in full (Table~\ref{tab:config}). Second, we never present numbers from different methodologies as a head-to-head: Table~\ref{tab:published} reproduces each vendor's own best self-reported LongMemEval figure, with the answer model that produced it, purely as the published landscape; the rows are not comparable to one another. We note one fact the table does support: Maximem Synap's figure was produced with a smaller answer model (gpt-5-mini) than the strongest competitor configurations, which is the clearest sign that the gain comes from the context layer, not the answer model.

\begin{table}[htbp]
\small
\centering
\caption{Published LongMemEval results (each vendor's own methodology; self-reported; not a controlled comparison)}
\label{tab:published}
\begin{tabularx}{\textwidth}{L{3.0cm} L{3.4cm} L{3.4cm} L{1.8cm} X}
\toprule
\textbf{System} & \textbf{LongMemEval (self-reported)} & \textbf{Answer model} & \textbf{Judge} & \textbf{Source} \\
\midrule
Maximem Synap & 92.0\% & gpt-5-mini & gpt-5-mini & Maximem (2026d) \\
SuperMemory & 85.2\% / 84.6\% / 81.6\% & Gemini-3 Pro / gpt-5 / gpt-4o & gpt-4o & SuperMemory (2026) research page \\
Zep & 71.2\% & gpt-4o & gpt-4o & Rasmussen et al.\ (2025) \\
Mem0 & not published & --- & --- & --- \\
Letta (MemGPT) & not published & --- & --- & --- \\
\bottomrule
\end{tabularx}
\end{table}

\subsection{Scope and limitations of this evaluation}
We want to be explicit about what these results and the system do and do not establish. Firstly, both benchmarks measure conversational recall and reasoning over recalled content. However, they do not measure latency under production load or token cost per task representing precision or robustness as retrieved context grows (``context rot''). Some of these are dimensions that production teams weigh heavily and on which this paper makes design arguments (Section 3) rather than benchmark claims. We deliberately report no latency numbers here and leave it for a follow up paper where we will propose a new benchmark that aims at plugging the gaps in existing benchmarks. Secondly, the limitations of the motivating study of Section 3.3 are already highlighted in that section. Thirdly, benchmark scores reflect a configuration and system combination; not a system alone in the abstract. Table~\ref{tab:config} is part of the result. Fourth, per-run artifacts are currently available on request rather than published; the harness and datasets are public.

Because the dimensions that matter to production teams are only partially captured by existing benchmarks, a comprehensive, reproducible benchmark for production context management, measuring accuracy, latency, token efficiency, and context-rot resistance together, is the subject of forthcoming work.

\section{Related Work}
We situate this paper in a field we survey rather than critique, mapping prior work onto the five primitives of Section 2. Characterizations of named systems are drawn from each system's own publications or documentation.

\textbf{Foundations:} The separation of memory contents from their use is an old idea in neural systems: differentiable neural computers coupled a controller network to an external, addressable memory (Graves et al., 2016). Cognitive-science-inspired surveys map human memory systems (sensory, working, episodic, semantic, procedural) onto AI analogues and emphasize that storage, retrieval, and forgetting are equally essential (Zihong He et al., 2024). Work framing forgetting and retention as facets of test-time memorization rather than deletion (Behrouz et al., 2025) informs how retention should be reasoned about. The in-context learning literature establishes that the shape of context, not merely its presence, drives model behavior (Dong et al., 2024).

\textbf{Memory systems (research):} MemGPT treats the LLM as an operating system managing a virtual-memory-like hierarchy of in-context and external storage, with the model itself paging information in and out (Packer et al., 2024); it primarily addresses the scoping and storage moments of the lifecycle, and has been productized as Letta, whose documentation describes stateful agents with developer-defined memory blocks, automatic compaction of older messages, and experimental ``sleep-time'' agents that process memory in the background between interactions, the closest published kin to the anticipating primitive (Letta docs, 2026). MIRIX structures agent memory into six typed components coordinated by a multi-agent framework that routes updates and retrieval (Wang \& Chen, 2025), addressing ingesting and scoping. Dynamic Cheatsheet maintains a self-updating external playbook that adapts at test time (Suzgun et al., 2025), and Agentic Context Engineering (ACE) extends the playbook idea with structured incremental updates that avoid the brevity bias of rewriting, documenting along the way the context-collapse failure central to Section 3.2 (Zhang et al., 2025); both sit at the intersection of ingesting and compacting. CAMELoT shows that a training-free associative-memory module can substitute for brute-force context scaling (Zexue He et al., 2024), a consolidation-flavored take on compacting.

\textbf{Memory systems (commercial):} Several commercial systems offer memory as an integrable layer; we characterize each in its own current terms (docs accessed June 12, 2026). Mem0 describes itself as ``a universal, self-improving memory layer for LLM applications,'' dynamically extracting and consolidating salient information from conversations, with a graph backend available in its open-source configuration (Chhikara et al., 2025; Mem0 docs, 2026); its published focus is the ingesting primitive. Zep positions itself as agent memory at enterprise scale, building a temporal ``Context Graph'' via Graphiti and assembling token-efficient context blocks from it (Zep/Graphiti docs, 2026); temporal fact validity is an ingestion-and-storage contribution, and its context assembly addresses scoping. SuperMemory describes itself as memory and context infrastructure for AI agents, combining fact-graph memory, pre-computed user profiles, and managed retrieval (SuperMemory docs, 2026). Cognee is an open-source AI memory platform that builds a knowledge graph from heterogeneous data behind a remember/recall/forget/improve interface (Cognee docs, 2026). Consumer assistants ship user-scoped memory natively (e.g., ChatGPT memory), which we note as the baseline most end users encounter. Mapped against Section 2, these systems concentrate on ingesting and storing memory, with partial scoping; per their own public material, none claims generative per-agent memory architecting, predictive pre-fetch as a product capability, or loss-validated compaction; the primitives, along with organization-scale scoping, where we believe the lifecycle framing adds the most.

\begin{table}[htbp]
\small
\centering
\caption{Systems $\times$ primitive coverage. \yes\ = stated primary focus per the system's own public documentation or paper; \phalf\ = partial or optional capability per the same sources; an empty cell means the dimension is not a stated focus of that system's public material, not a claim that it is absent. All sources accessed June 12, 2026; row-by-row citations in the supplementary notes.}
\label{tab:coverage}
\begin{tabularx}{\textwidth}{L{3.4cm} c c c c X}
\toprule
\textbf{System} & \textbf{Architecting} & \textbf{Ingesting} & \textbf{Scoping} & \textbf{Anticipating} & \textbf{Compacting \& Consolidation} \\
\midrule
MemGPT / Letta & ---\textsuperscript{a} & \phalf & \yes\textsuperscript{b} & \phalf\textsuperscript{c} & \yes \\
MIRIX & & \yes & \phalf & & \phalf \\
ACE / Dynamic Cheatsheet & & \phalf\textsuperscript{d} & & & \phalf\textsuperscript{e} \\
Mem0 & & \yes & \phalf & & \phalf \\
Zep / Graphiti & ---\textsuperscript{a} & \yes & \yes & & \phalf \\
SuperMemory & & \yes & \phalf & \phalf\textsuperscript{f} & \phalf \\
Cognee & ---\textsuperscript{a} & \yes & \phalf & & \phalf \\
Maximem Synap (this paper) & \yes & \yes & \yes & \yes & \yes \\
\bottomrule
\end{tabularx}
\end{table}

\noindent\footnotesize
\textsuperscript{a} Several systems support developer-defined or data-emergent structure (Letta's developer-defined memory blocks; Graphiti's ``learned ontology''; Cognee's ontology support). We mark architecting only where a system generates a per-agent memory architecture from a description of the agent's purpose; structure that is hand-defined or that emerges post-hoc from ingested data is a different capability.
\textsuperscript{b} On the token-budgeting reading: context-window management is MemGPT's founding thesis and Letta documents an explicit context hierarchy with size budgets. Hierarchical organizational scoping (user vs.\ organization vs.\ platform) is not a stated focus.
\textsuperscript{c} Letta's sleep-time agents process memory in the background between interactions; their own docs mark the feature experimental. Background pre-computation is adjacent to, though not the same as, query-predictive pre-fetch.
\textsuperscript{d} ACE and Dynamic Cheatsheet extract strategies from the agent's own execution experience rather than from conversations or documents.
\textsuperscript{e} ACE addresses information loss in context rewriting by avoiding compression (incremental structured updates); it does not perform validated summarization.
\textsuperscript{f} SuperMemory's user profiles are pre-computed standing context (``no search needed''), not query-predictive pre-fetch.
\textsuperscript{g} The source post states 97.8\%; the correct arithmetic for 38 usable of 10,134 is a 99.6\% junk rate. We report the raw counts and the corrected percentage.
\normalsize

\textbf{Retrieval:} Retrieval-augmented generation is the substrate (Lewis et al., 2020). Graph-aware variants extend it toward relational and multi-document reasoning: GraphRAG builds community-summarized entity graphs for corpus-level questions (Edge et al., 2024), and HippoRAG uses a hippocampus-inspired index over a knowledge graph for multi-hop retrieval (Guti\'errez et al., 2024). RAPTOR retrieves over a recursively summarized tree (Sarthi et al., 2024). ReMindRAG memorizes LLM traversal decisions in knowledge-graph edge embeddings so that similar queries can replay paths without fresh per-hop LLM calls (Hu et al., 2025), and is, with GraphRAG and HippoRAG, the published prior art closest to the graph-aware retrieval of Section 4, we cite it as such.

\textbf{Long context:} A complementary line shows that enlarging the context window is not free: models degrade on information in the middle of long contexts (Liu et al., 2024), and streaming attention (Xiao et al., 2024) and KV-cache compression (Li et al., 2024) attack the serving cost of long contexts at the attention level. ACM is orthogonal; it decides what should be in the window in the first place, whatever the window's size.

\section{Future Directions: Decision-Level and Organization-Scale Context}
The lifecycle of Section 2 manages conversational and organizational context today. The frontier is decision-level context: capturing not just what happened but why an organization decided what it did, so that agents can reason with institutional judgment rather than only institutional facts. This is the ``context graph'' opportunity now drawing attention in the investment community, where the broader opportunity has been estimated to be in the trillions (Gupta \& Garg, 2025), and we have published an extended treatment of it (Dadhich, 2026b).

We are deliberately measured about it, because the hard problems are real and largely unsolved. Most decisions are implicit and never recorded; recorded rationales are often post-hoc justifications rather than true traces; tying decisions to outcomes requires causal attribution that even humans struggle with; and knowing when a past decision has been superseded is its own problem. There are equally hard technical problems: extracting decision traces across organizational systems, intra-organization canonicalization (the same document called ``PR FAQ'' by one team and ``6-pager'' by another), temporal grounding, what to store and where, and the security consequences of co-locating previously siloed knowledge in one logical system. We raise these challenges not because we have solved all of them (although we have solved most in the reference system), but because listing them is how the category can attempt to solve it together and mature the category itself. A context-management lifecycle that today manages facts across users and organizations is the necessary substrate on which decision-level context can eventually be built.

\section{Conclusion}
Agents do not primarily fail to deliver on their promise due to lack of intelligence but they rather fail due to the lack of managed context. Treating that as a memory-storage problem under-describes it. We have argued that it is a lifecycle (architecting, ingesting, scoping, anticipating, compacting \& consolidation) that operates across an organizational scope hierarchy, and that managing it as a system is an economic necessity, not a convenience. It is the gap between quadratic and linear cost, and between confident wrong answers and validated ones. The next generation of agent context infrastructure will not just offer a location to store history but will be a system that manages the full context lifecycle. We have described one such system, and pointed to the decision-level frontier the category is heading toward. The contest ahead is not over who stores the most data; it is over who manages context the best.

\section*{Acknowledgments}
The author thanks Shreyansh Singh Gautam and Anish Yadav from the Maximem team for their work on the implementation of Maximem Synap as well as review feedback on this paper. The author also thanks Varun Gupta and other reviewers who provided very detailed and high-quality review feedback that materially improved this paper.

\section*{References}
{\small
\refitem{Behrouz, A., Razaviyayn, M., Zhong, P., \& Mirrokni, V. (2025). It's All Connected: A Journey Through Test-Time Memorization, Attentional Bias, Retention, and Online Optimization (Miras). arXiv:2504.13173v1. ICLR 2026 (poster).}
\refitem{Chhikara, P., Khant, D., Aryan, S., Singh, T., \& Yadav, D. (2025). Mem0: Building Production-Ready AI Agents with Scalable Long-Term Memory. arXiv:2504.19413v1.}
\refitem{Cognee (2026). Cognee documentation, docs.cognee.ai; GitHub: topoteretes/cognee. Accessed June 12, 2026.}
\refitem{Dadhich, G. (2026a). File vs Vector for RAG. Maximem blog, January 15, 2026. \url{https://www.maximem.ai/blog/file-rag-vs-vector-rag} (data: github.com/maximem-ai/file-vs-vector-study-results)}
\refitem{Dadhich, G. (2026b). Context Graphs: The Trillion Dollar Elephant. Medium, January 22, 2026.}
\refitem{Dong, Q., et al.\ (2024). A Survey on In-context Learning. arXiv:2301.00234v6. EMNLP 2024.}
\refitem{Edge, D., Trinh, H., Cheng, N., et al.\ (2024). From Local to Global: A Graph RAG Approach to Query-Focused Summarization. arXiv:2404.16130v2.}
\refitem{Graves, A., et al.\ (2016). Hybrid computing using a neural network with dynamic external memory. Nature, 538:471--476.}
\refitem{Gupta, J., \& Garg, A. (2025). AI's trillion-dollar opportunity: Context graphs. Foundation Capital, December 22, 2025. \url{https://foundationcapital.com/ideas/context-graphs-ais-trillion-dollar-opportunity}}
\refitem{Guti\'errez, B. J., et al.\ (2024). HippoRAG: Neurobiologically Inspired Long-Term Memory for Large Language Models. arXiv:2405.14831v3. NeurIPS 2024.}
\refitem{He, Zexue, Karlinsky, L., Kim, D., McAuley, J., Krotov, D., \& Feris, R. (2024). CAMELoT: Towards Large Language Models with Training-Free Consolidated Associative Memory. arXiv:2402.13449v1.}
\refitem{He, Zihong, Lin, W., Zheng, H., et al.\ (2024). Human-inspired Perspectives: A Survey on AI Long-term Memory. arXiv:2411.00489v2.}
\refitem{Hu, Y., Zhu, J., Tang, L., \& Huang, C. (2025). ReMindRAG: Low-Cost LLM-Guided Knowledge Graph Traversal for Efficient RAG. arXiv:2510.13193v2. NeurIPS 2025.}
\refitem{Letta (2026). Letta documentation, docs.letta.com. Accessed June 12, 2026.}
\refitem{Lewis, P., et al.\ (2020). Retrieval-Augmented Generation for Knowledge-Intensive NLP Tasks. NeurIPS 2020. arXiv:2005.11401.}
\refitem{Li, Y., Huang, Y., Yang, B., et al.\ (2024). SnapKV: LLM Knows What You are Looking for Before Generation. arXiv:2404.14469v2. NeurIPS 2024.}
\refitem{Liu, N. F., et al.\ (2024). Lost in the Middle: How Language Models Use Long Contexts. TACL 12:157--173. arXiv:2307.03172v3.}
\refitem{Maharana, A., Lee, D.-H., Tulyakov, S., Bansal, M., Barbieri, F., \& Fang, Y. (2024). Evaluating Very Long-Term Conversational Memory of LLM Agents (LoCoMo). ACL 2024, pp.\ 13851--13870. arXiv:2402.17753v1.}
\refitem{Maximem (2026a). How Synap Works Under the Hood. maximem.ai blog, April 11, 2026. \url{https://www.maximem.ai/blog/how-maximem-synap-works}}
\refitem{Maximem (2026c). Maximem Synap Updates: Higher Scores, 17 Integrations, and a Live Playground. maximem.ai blog, May 27, 2026. \url{https://www.maximem.ai/blog/maximem-synap-updates-higher-benchmark-scores-and-more}}
\refitem{Maximem (2026d). Synap: Agent Memory Benchmark Results. GitHub: maximem-ai/eval\_benchmark\_runs\_output (README, RESULTS.md, METHODOLOGY.md; CITATION.cff; CC BY 4.0). Accessed July 9, 2026.}
\refitem{Maximem (2026e). Memory and Context Eval Harness. GitHub: maximem-ai/memory\_and\_context\_eval\_harness.}
\refitem{McKinsey \& Company (2025). The State of AI in 2025: Agents, Innovation, and Transformation. QuantumBlack by McKinsey, November 2025. \url{https://www.mckinsey.com/capabilities/quantumblack/our-insights/the-state-of-ai}}
\refitem{Mem0 (2026). Mem0 documentation, docs.mem0.ai. Accessed June 12, 2026.}
\refitem{Packer, C., et al.\ (2024). MemGPT: Towards LLMs as Operating Systems. arXiv:2310.08560v2.}
\refitem{Rasmussen, P., et al.\ (2025). Zep: A Temporal Knowledge Graph Architecture for Agent Memory. arXiv:2501.13956.}
\refitem{Sarthi, P., et al.\ (2024). RAPTOR: Recursive Abstractive Processing for Tree-Organized Retrieval. ICLR 2024. arXiv:2401.18059v1.}
\refitem{SuperMemory (2026). SuperMemory documentation, supermemory.ai/docs. Accessed June 12, 2026.}
\refitem{SuperMemory (2026). Research: LongMemEval results. supermemory.ai/research. Accessed July 9, 2026 (via Maximem, 2026d).}
\refitem{Suzgun, M., Yuksekgonul, M., Bianchi, F., Jurafsky, D., \& Zou, J. (2025). Dynamic Cheatsheet: Test-Time Learning with Adaptive Memory. arXiv:2504.07952v1.}
\refitem{Wang, Y., \& Chen, X. (2025). MIRIX: Multi-Agent Memory System for LLM-Based Agents. arXiv:2507.07957v1.}
\refitem{Wu, D., Wang, H., Yu, W., Zhang, Y., Chang, K.-W., \& Yu, D. (2024). LongMemEval: Benchmarking Chat Assistants on Long-Term Interactive Memory. arXiv:2410.10813v2. ICLR 2025.}
\refitem{Xiao, G., et al.\ (2024). Efficient Streaming Language Models with Attention Sinks. arXiv:2309.17453v4. ICLR 2024.}
\refitem{Zep / Graphiti (2026). Zep documentation, help.getzep.com; Graphiti GitHub: getzep/graphiti. Accessed June 12, 2026.}
\refitem{Zhang, Q., Hu, C., Upasani, S., et al.\ (2025). Agentic Context Engineering: Evolving Contexts for Self-Improving Language Models. arXiv:2510.04618v3. ICLR 2026.}
}

\appendix
\section{Cost-model derivation}
Let each turn add $t$ tokens and the conversation run $n$ turns. Full-append re-sends the entire history each turn, so input tokens at turn $k$ are $k\!\cdot\!t$ and the cumulative total is
\[
C_{\text{append}} \;=\; \sum_{k=1}^{n} k\,t \;=\; t\,\frac{n(n+1)}{2}.
\]
A bounded-budget system holds input to $W$ tokens per turn: $C_{\text{bounded}} = nW$. The cost multiple is
\[
R(n) \;=\; \frac{C_{\text{append}}}{C_{\text{bounded}}} \;=\; \frac{t\,(n+1)}{2W},
\]
linear in $n$. Illustrative values ($t = 500$, $W = 4{,}000$):

\begin{table}[htbp]
\centering
\begin{tabular}{l r r r}
\toprule
\textbf{Turns $n$} & \textbf{Full-append (tokens)} & \textbf{Bounded (tokens)} & \textbf{Multiple} \\
\midrule
50 & 637,500 & 200,000 & 3.2$\times$ \\
100 & 2,525,000 & 400,000 & 6.3$\times$ \\
200 & 10,050,000 & 800,000 & 12.6$\times$ \\
500 & 62,625,000 & 2,000,000 & 31.3$\times$ \\
\bottomrule
\end{tabular}
\end{table}

These are illustrative, not measured; they assume constant per-turn token addition and ignore caching discounts, which shift the constants but not the asymptotics. Per-call attention compute scales quadratically in sequence length per turn, making cumulative compute roughly cubic in $n$ for full-append; we lead with the billing result because it is the cleaner claim.

The frontier argument of Section 3.2 follows: full-append buys fidelity at quadratic cost; crude summarization buys linear cost by surrendering fidelity (Zhang et al., 2025, document a single-step compression of 18,282 $\rightarrow$ 122 tokens dropping accuracy from 66.7\% to 57.1\%, below the no-context baseline); validated compaction targets linear cost with checked fidelity.

\section{Retrieval-study methodology (motivating study, Section 3.3)}
\setcounter{table}{0}
\renewcommand{\thetable}{B\arabic{table}}
\textbf{Setup.} Five public datasets spanning distinct retrieval regimes: CodeXGLUE (natural language $\rightarrow$ code), MS MARCO (web passages), SQuAD (factoid QA), HotpotQA (multi-hop), SciQ (science). Five separate corpora: 10,000 documents per dataset (50,000 in aggregate, indexed per-dataset, not as one combined corpus), with 1,000 queries per dataset (5,000 total), scored by MRR@10 against each dataset's gold annotations. Substrates: Tantivy 0.22.0 (default text analyzer) and ChromaDB ($\geq$0.4.0, default index) with all-MiniLM-L6-v2 embeddings (384-dim). No chunking: documents were indexed whole. Hardware: Apple M4 MacBook Pro (10 cores, 16 GB RAM). Runs executed January 14, 2026. The public write-up (Dadhich, 2026a) reported these findings qualitatively; the tables below publish the numbers for the first time. The per-dataset scores and configuration are released at maximem-ai/file-vs-vector-study-results.

\begin{table}[htbp]
\small
\centering
\caption{MRR@10 by dataset, keyword vs.\ vector retrieval (10,000 docs / 1,000 queries per dataset)}
\label{tab:b1}
\begin{tabular}{l c c l}
\toprule
\textbf{Dataset (regime)} & \textbf{Keyword (Tantivy)} & \textbf{Vector (Chroma)} & \textbf{Leader} \\
\midrule
CodeXGLUE (NL $\rightarrow$ code) & 0.290 & 0.914 & Vector, decisively \\
MS MARCO (web queries) & 0.404 & 0.523 & Vector \\
SQuAD (factoid QA) & 0.605 & 0.614 & Parity \\
HotpotQA (multi-hop) & 0.549 & 0.495 & Keyword, narrowly \\
SciQ (science) & 0.815 & 0.614 & Keyword, decisively \\
\bottomrule
\end{tabular}
\end{table}

\begin{table}[htbp]
\small
\centering
\caption{The vector tax: wall-clock indexing time per 10,000-document corpus}
\label{tab:b2}
\begin{tabular}{l c c c}
\toprule
\textbf{Corpus} & \textbf{Keyword indexing} & \textbf{Embedding + vector indexing} & \textbf{Multiple} \\
\midrule
CodeXGLUE & 0.45 s & 43.1 s & 97$\times$ \\
MS MARCO & 0.42 s & 26.3 s & 63$\times$ \\
SQuAD & 0.41 s & 35.9 s & 88$\times$ \\
HotpotQA & 0.39 s & 29.5 s & 75$\times$ \\
SciQ & 0.44 s & 26.7 s & 60$\times$ \\
\bottomrule
\end{tabular}
\end{table}

\textbf{Limitations} (stated in full; these are why Section 3.3 presents the study as motivation, not benchmark):
\begin{enumerate}[leftmargin=1.4em,itemsep=1pt,topsep=2pt]
\item Single operator; one keyword engine against one vector store. Results characterize this configuration, not retrieval methods in the abstract.
\item No chunking. all-MiniLM-L6-v2 truncates input beyond its maximum sequence length, so on long documents the vector index effectively embedded only the head of each document. This biases long-text comparisons against the vector side; with chunking, vector results would likely improve on those corpora. CodeXGLUE's natural-language-to-code snippets, by contrast, are short enough to fit the model's window, so truncation does not affect that corpus and its 0.914 result stands.
\item Small per-corpus scale. 10,000-document indexes do not exercise production-scale vector-search constraints.
\item Single-gold-target scoring. HotpotQA was scored against one supporting document per question; the evaluation design cannot measure multi-document sufficiency. Section 3.3 discusses why this limitation is itself informative.
\item Per-query traces were not retained by the evaluation runner; the accompanying release contains per-dataset scores and the evaluation code, not per-query outputs.
\item MS MARCO's license restricts redistribution of passage text; the release excludes underlying passages for that corpus.
\item The study compared the two retrieval substrates in isolation; it did not evaluate a fused hybrid pipeline (for example, reciprocal-rank fusion of lexical and semantic results) or a second-stage cross-encoder reranker, both of which would be expected to raise the vector and hybrid numbers. The claim is about the necessity of combining signals, not a measurement of the best achievable combined system.
\end{enumerate}

The conclusions Section 3.3 draws, regime-dependence of retrieval method (B1), the indexing-time asymmetry (B2), and the structural blindness of hit-based scoring to sufficiency, are each robust to caveats 1--3, and caveat 4 is the basis of the sufficiency argument rather than a threat to it.

\section{Reproducibility statement}
For every result in Section 6 we state: the dataset and split with question counts, the answer and judge models, the full retrieval configuration, the harness commit SHA, the run date, and what we release publicly (harness code, configs, and raw per-question outputs). If a result cannot be re-run by a third party at publication time, we say so directly rather than imply otherwise. Any competitor number follows the two rules in Section 6.2: identical protocol, or clearly labeled as self-reported, and never mixed in one table.

\section{Data use and privacy}
Maximem Synap processes conversational data that can include personal information. At the policy level: memory is tenant-isolated by construction (storage- and query-layer enforcement, Section 4); customers control retention through the generated architecture's guardrails; and personally identifying information is subject to extraction-time handling policies configured per agent. This paper reports no customer data; benchmark results use public datasets.

\end{document}